\documentclass[10pt]{article}

\usepackage[margin=0.5in]{geometry}

\usepackage{amsmath}
\usepackage{amsthm}
\usepackage{amssymb}

\usepackage[utf8]{inputenc}
\usepackage{hyperref}

\usepackage{graphicx, color}
\usepackage{tabularx}

\usepackage{booktabs}

\title{Data-Efficient Sleep Staging with Synthetic Time Series Pretraining}
\author{Niklas Grieger$^{1,2,3,*}$ \and Siamak Mehrkanoon$^2$ \and Stephan Bialonski$^{1,3,*}$}

\date{
	$^1$Department of Medical Engineering and Technomathematics, FH Aachen University of Applied Sciences, 52428 Jülich, Germany\\
	$^2$Department of Information and Computing Sciences, Utrecht University, Utrecht, The Netherlands\\
	$^3$Institute for Data-Driven Technologies, FH Aachen University of Applied Sciences, 52428 Jülich, Germany\\
    $^*$\texttt{grieger@fh-aachen.de, bialonski@fh-aachen.de}
    \\[2ex]%
}

\begin{document}
\maketitle

\begin{abstract}
    Analyzing electroencephalographic (EEG) time series can be challenging, especially with deep neural networks, due to the large variability among human subjects and often small datasets.
    To address these challenges, various strategies, such as self-supervised learning, have been suggested, but they typically rely on extensive empirical datasets.
    Inspired by recent advances in computer vision, we propose a pretraining task termed ``frequency pretraining'' to pretrain a neural network for sleep staging by predicting the frequency content of randomly generated synthetic time series.
    Our experiments demonstrate that our method surpasses fully supervised learning in scenarios with limited data and few subjects, and matches its performance in regimes with many subjects.
    Furthermore, our results underline the relevance of frequency information for sleep stage scoring, while also demonstrating that deep neural networks utilize information beyond frequencies to enhance sleep staging performance, which is consistent with previous research.
    We anticipate that our approach will be advantageous across a broad spectrum of applications where EEG data is limited or derived from a small number of subjects, including the domain of brain-computer interfaces.
\end{abstract}

\section{Introduction}

Deep neural networks have achieved significant advances in analyzing electroencephalographic (EEG) time series~\cite{Roy2019}, ranging from brain-computer interfaces~\cite{Ko2021} to the intricacies of sleep stage scoring~\cite{Phan2022,Fiorillo2019}.
Such successes are attributed to the ability of deep neural networks, as universal function approximators, to learn properties (features) from patient data that are difficult for humans to conceptualize and define.
However, training neural networks requires large and diverse datasets that capture the considerable variety between individual subjects and their medical conditions (subject heterogeneity).
Creating such datasets is challenging due to the typically limited amount of data per subject (data scarcity) and diverse measurement protocols used in different clinics, which can introduce additional variability in the data.
Furthermore, acquiring large datasets is often expensive, complicated, or even intractable due to strict privacy policies and ethical guidelines.
This hinders the advancement of deep neural networks for widespread application in real-world medical settings.

Efforts to mitigate the scarcity of large datasets have primarily followed two paths: (1) the development of network architectures that incorporate constraints mirroring the data's intrinsic characteristics, such as symmetries~\cite{Bronstein2021}, and (2) enhancing model performance with additional or cross-domain data to learn effective priors.
Pertaining to the first path, a common feature in time series processing networks is the use of convolutional layers.
These layers are designed to be translation-equivariant~\cite{Goodfellow2016}, which ensures that a temporal shift in the input only affects the output by the same shift.
This characteristic enables consistent network responses to temporal patterns, regardless of their temporal location, while reducing the number of model parameters compared to architectures lacking such constraints.
For the second path, a variety of strategies have been proposed to learn useful priors from data.
One approach is data augmentation, in which time series are transformed while preserving their annotations (labels) to artificially expand the dataset~\cite{Lashgari2020,He2021}.
Deep neural networks trained on such augmented datasets implicitly learn to become invariant under these transformations, which can lead to better out-of-sample prediction performance.
Another strategy is transfer learning~\cite{Ebbehoj2022}, a two-step process in which neural networks are trained on one task using a large dataset (pretraining step~\cite{Qiu2023}) and then adapted to learn the actual task of interest using another (usually much smaller) dataset (fine-tuning step).
A variant of this idea is self-supervised learning~\cite{Liu2023,Banville2021}, which allows neural networks to be pretrained on large and heterogeneous datasets without explicitly labeled examples.
Finally, generative models such as VAEs, GANs, and diffusion models can be used to sample new time series to extend existing datasets~\cite{Habashi2023,Carrle2023,You2025}.
Such generative models approximate a data distribution and require large heterogeneous datasets for training.
While all of these approaches have been demonstrated to be able to improve the performance of neural networks, they still rely on large empirical datasets for training.

Recent advances in computer vision have demonstrated that it is possible to learn effective priors exclusively from synthetic images, which has the potential to significantly reduce the need for large empirical datasets~\cite{Baradad2021,Kataoka2022}.
Synthetic images for image classification tasks were generated by simple random processes, such as iterated function systems to produce fractals~\cite{Kataoka2022} or random placement of geometric objects to cover an image canvas~\cite{Baradad2021}.
Deep neural networks pretrained on such data were demonstrated to learn useful priors for image classification tasks, yielding competitive performance comparable to pretraining on natural images on various benchmarks~\cite{Kataoka2022}.
This remarkable finding highlights the potential of synthetic datasets that can be generated without much computational resources and, theoretically, in unlimited amounts.

Inspired by these advances, we hypothesize that pretraining exclusively on synthetic time series data generated from simple random processes can also yield effective priors for sleep staging.
Given the importance of frequencies for sleep stage scoring and other EEG-based applications~\cite{Berry2020,Motamedi2014}, we introduce a pretraining method called ``frequency pretraining'' (FPT) that centers on generating synthetic time series data with specific frequency content.
During pretraining, deep neural networks learn to accurately predict the frequencies present in these synthetic time series.
Despite the deliberate simplicity of our synthetic data generation process and the inherent domain shift between synthetic and EEG data, we observe that FPT allows a deep neural network to detect sleep stages more accurately than fully supervised training when few samples (few-sample regime) or data from few subjects (few-subject regime) are available for fine-tuning.
The success of our method underscores the essential role of frequency content in enabling neural networks to accurately and reliably discern sleep stages.
We consider pretraining techniques leveraging synthetic data, like the one we propose, as a promising area of research, offering the potential to develop models in sleep medicine and neuroscience that are particularly suited for scenarios involving small datasets.
To facilitate testing and further advancements, we make the source code of our method publicly available~\cite{Grieger2024}.

\subsection*{Contributions}

\begin{itemize}
    \item Novel synthetic pretraining approach: We introduce ``frequency pretraining'' (FPT), a pretraining approach using synthetic time series with random frequency content, eliminating the need for empirical EEG data during pretraining.
    \item Demonstrated data efficiency: We demonstrate superior sleep staging performance of FPT in few-sample and few-subject regimes across three datasets, with comparable results to fully supervised methods when data is abundant.
    \item Analysis of frequency-based priors: We evaluate the role of frequency information in our pretraining task and how synthetic sample diversity affects fine-tuning.
    \item Comparison with self-supervised methods: We benchmark our approach against established self-supervised learning methods, demonstrating that synthetic data pretraining can achieve comparable results without requiring EEG data for pretraining.
\end{itemize}

\section{Results}

\begin{figure}[h]
    \centering
    \includegraphics[width=\textwidth]{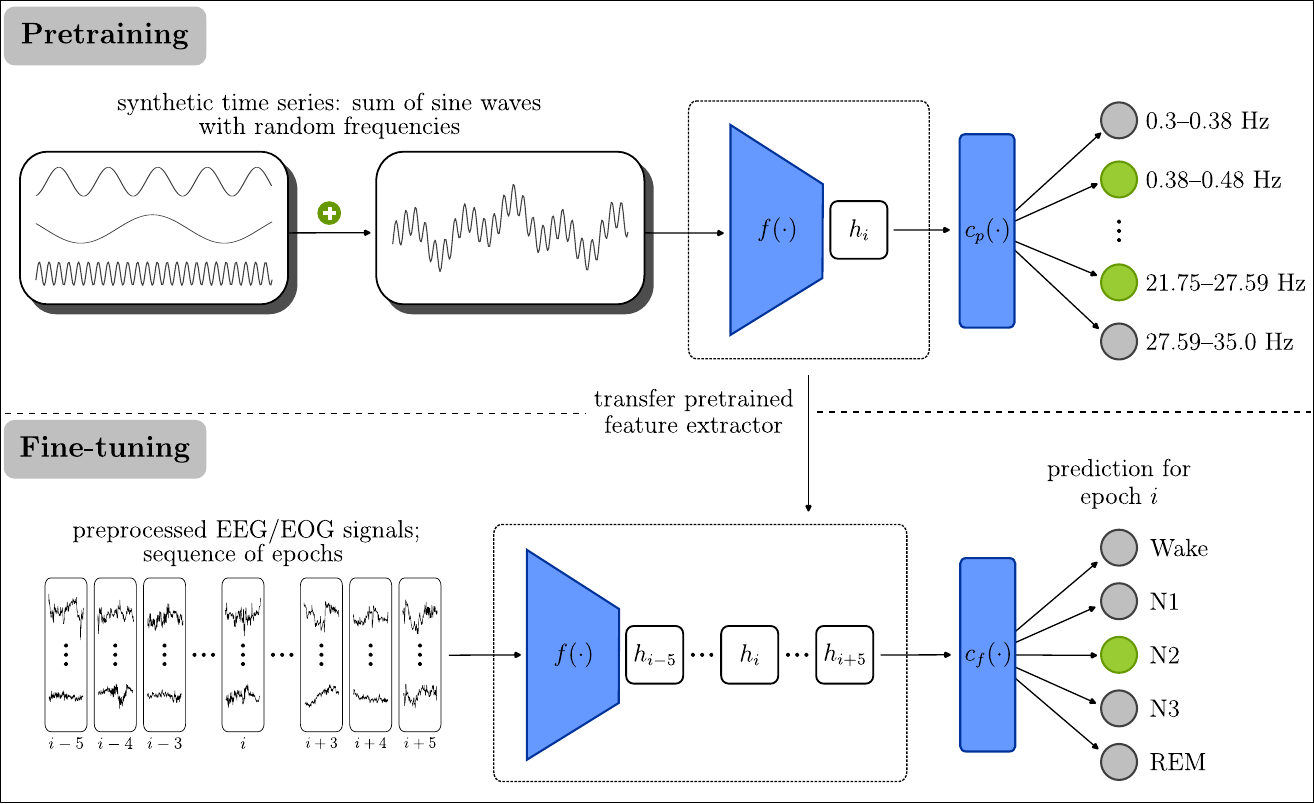}
    \caption{
        The training process consists of a \emph{pretraining} and a \emph{fine-tuning} phase.
        In the pretraining phase (``frequency pretraining''), we generated synthetic time series signals by summing sine waves with random frequencies.
        These synthetic signals were used to train a deep neural network consisting of a feature extractor $f$ and a classifier $c_p$ to predict the frequencies present in the signals (multi-label classification problem).
        In the fine-tuning phase, the pretrained feature extractor $f$ produced features $h_i$ from individual EEG and EOG epochs.
        The features of a sequence of these epochs were then used by a classifier $c_f$ that was trained to predict the sleep stage of the middle epoch in the sequence (multi-class classification problem).
    }
    \label{fig:pretraining-scheme}
\end{figure}

Our approach, illustrated in Figure~\ref{fig:pretraining-scheme} and detailed in Section~\ref{sec:methods}, is based on a two-phase training process that combines pretraining on synthetic time series with fine-tuning on clinical sleep data.
During the pretraining phase, we generate synthetic time series composed of sine waves with random frequencies drawn from predefined frequency ranges (frequency bins).
These synthetic signals are then used to train a deep neural network whose feature extractor, $f$, learns to extract useful features, while the classifier, $c_p$, learns to predict the frequency bins from which the frequencies were drawn to generate the synthetic time series.
After pretraining, the feature extractor is transferred to the fine-tuning phase, where it is applied to real EEG and electrooculography (EOG) data to classify sleep stages.
In this phase, the model processes sequences of eleven consecutive sleep epochs, indexed $i-5$ to $i+5$, with the feature extractor producing features $h_i$ from each epoch.
Another classifier $c_f$ then aggregates these features to predict the sleep stage (Wake, N1, N2, N3, REM) for the central sleep epoch $i$.

We evaluated our approach on three publicly available datasets, \emph{DODO/H}, \emph{Sleep-EDFx}, and~\emph{ISRUC}, which provided data from 276 subjects, including both healthy individuals and those with various medical conditions (see Section~\ref{sec:sleep-staging-data}).
For sleep staging performance, we tracked the Macro-F1 score, with higher values indicating better classification accuracy across sleep stages.
During the pretraining phase, we assessed the model's ability to predict frequency bins using the Hamming metric and the accuracy (see Section~\ref{sec:methods}).

\phantomsection
\subsection{Training Configurations}
\label{sec:training-setups}

We compared the performance of pretrained models against the performance of non-pretrained models in scenarios with varying amounts of training data.
In particular, we studied the performance of our approach in few-sample and few-subject regimes, where the greatest benefit was expected.
Furthermore, we analyzed the priors that the model learned during pretraining and the role of frequency information in the learned features.
Finally, we investigated whether these features could be further improved by fine-tuning the feature extractor.
To enable these investigations, we created four training configurations.

\textbf{Fully Supervised.}
The Fully Supervised training configuration is similar to many existing deep learning approaches for sleep staging~\cite{Phan2022} and served as a baseline to compare our pretrained models against.
In this configuration, we skipped the pretraining step and trained (fine-tuned) the feature extractor $f$ and classifier $c_f$ from scratch using sleep staging data.

\textbf{Fixed Feature Extractor.}
We employed the Fixed Feature Extractor configuration to investigate the relevance of the features generated by the pretrained feature extractor for sleep staging.
After pretraining the feature extractor $f$ on synthetic data, we kept its model weights and BatchNorm statistics fixed and only fine-tuned the sleep staging classifier $c_f$ on sleep data.

\textbf{Fine-Tuned Feature Extractor.}
With this training configuration, we studied (i) how model performance changes when the pretrained feature extractor is allowed to change during fine-tuning and (ii) whether the priors learned during pretraining can prevent overfitting in few-sample or few-subject regimes.
As in the previous configuration, we first pretrained the feature extractor $f$ on synthetic data, but then fine-tuned both the feature extractor and the classifier $c_f$ on sleep data without keeping any model weights fixed.
Consequently, this configuration is similar to the Fully Supervised configuration with the key distinction that the feature extractor is initialized with pretrained weights.

\textbf{Untrained Feature Extractor.}
The Untrained Feature Extractor configuration served as a baseline to study whether our pretraining scheme produces priors that are superior to random weights for sleep staging.
We randomly initialized the feature extractor $f$ using Kaiming normal initialization~\cite{He2015} and then kept its weights fixed while fine-tuning the classifier $c_f$.
This approach mirrors the Fixed Feature Extractor configuration, but with a random feature extractor instead of a pretrained one.

\subsection{Data Efficiency}
\label{sec:data-efficiency}

To assess the data efficiency of our pretraining method, we compare the performance of the different training configurations when fine-tuned with a reduced amount of training data (low-data regime) or the full training data (high-data regime).
We investigated the low-data regime by first pretraining models with the full synthetic dataset.
Depending on the training configuration, we then fine-tuned the pretrained or randomly initialized models with the data of 50 randomly sampled sleep staging samples from one subject.
The sampling procedure selected sleep staging data without class stratification, and each sleep stage was represented at least once in the reduced datasets.
In the high-data regime, we followed the same procedure but fine-tuned models with the full training data instead of a reduced amount of data.
We repeated all experiments three times in a 5-fold cross-validation scheme, resulting in 15 training runs for each configuration and dataset.
This approach allowed us to estimate the spread of the macro F1 scores when using different model initializations, sleep staging samples, and data folds for training and testing (see Section~\ref{sec:methods}).

\begin{figure}
    \centering
    \includegraphics[width=\textwidth]{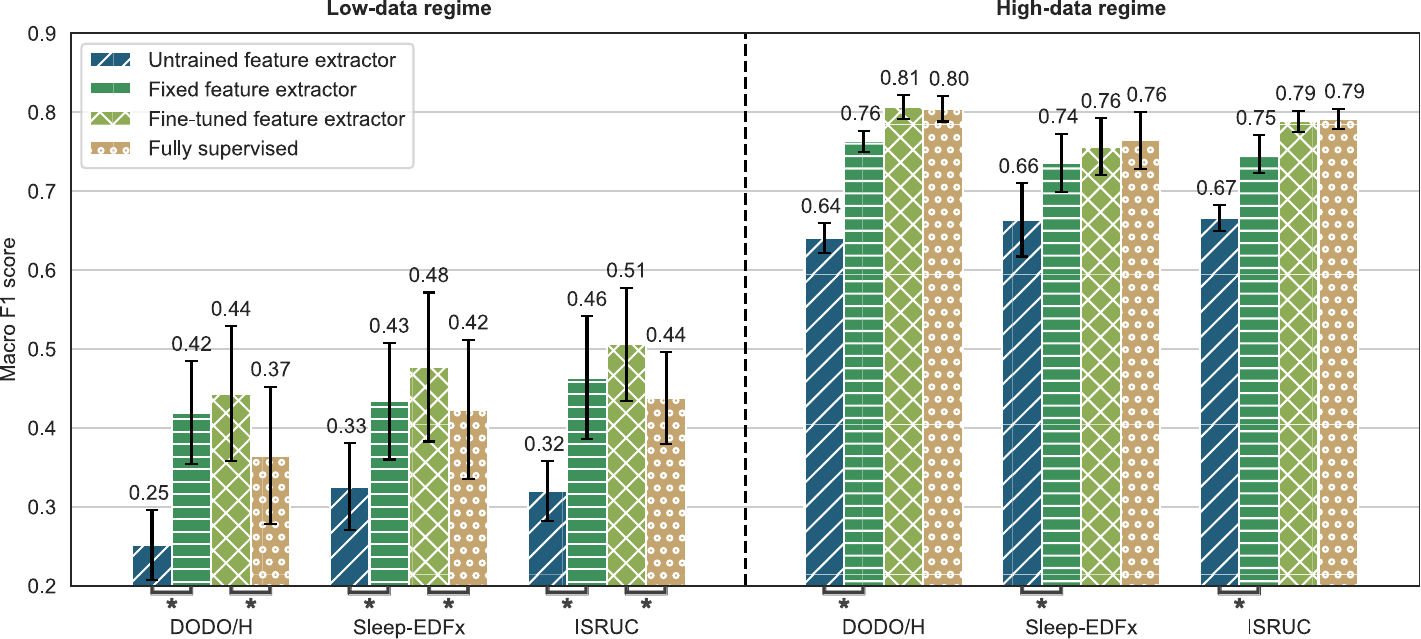}
    \caption{
            Average macro F1 scores achieved by the different training configurations when fine-tuned with a small set of training data (left side) or a large set (full set) of training data (right side) from one of the three datasets DODO/H, Sleep-EDFx, and ISRUC.
            The bars indicate the mean of the macro F1 scores measured on the test sets and averaged over 15 trainings (3 repetitions of a 5-fold cross-validation).
            Error bars show the standard deviation of the macro F1 scores, and~* symbols indicate significant differences between Untrained and Fixed Feature Extractor or between Fine-Tuned Feature Extractor and Fully-Supervised training configuration $(p<0.01)$ according to a one-sided Wilcoxon signed-rank test.
    }
    \label{fig:subj-vs-mf1}
\end{figure}

In the low-data regime, we observed that models pretrained with synthetic data outperformed models trained from scratch in sleep stage classification (see Figure~\ref{fig:subj-vs-mf1}).
The performance gap between pretrained and non-pretrained models was most pronounced when comparing the fine-tuned feature extractor to the fully supervised configuration, with the former achieving average macro F1 scores that were 0.06--0.07 higher than those of the latter across datasets.
When removing the fine-tuning of the pretrained features (fixed feature extractor configuration), our method still yielded macro F1 scores that were 0.01--0.05 higher than those of the fully supervised configuration.
Comparing the fixed feature extractor to the untrained feature extractor configuration further highlights the importance of the learned features.
Pretraining the feature extractor with synthetic data improved the macro F1 scores by 0.10--0.17 compared to a random initialization.
While the macro F1 scores of the training configurations varied between datasets, the general trends observed in the low-data regime were consistent across all three datasets.

In the high-data regime, our pretrained models were on par with fully supervised trained models (\emph{p} < 0.05 for a paired TOST test with a margin of $\pm 0.01$ on the DODO/H and ISRUC datasets) and achieved competitive performance in sleep stage classification (see Figure~\ref{fig:subj-vs-mf1}).
The fine-tuned feature extractor configuration achieved average macro F1 scores of 0.76--0.81 across datasets, comparable to the macro F1 scores of 0.76--0.80 achieved by the fully supervised configuration.
As in the low-data regime, we observed that fine-tuning the pretrained features was beneficial, as the macro F1 scores achieved by the fine-tuned feature extractor configuration were 0.02--0.05 higher than those of the fixed feature extractor configuration.
The performance gap between the untrained feature extractor and the fixed feature extractor remained substantial even in the high-data regime, with the former achieving average macro F1 scores that were 0.08--0.12 lower than those of the latter.

\subsection{Impact of Subject Diversity and Number of Training Samples}

We further explored the data efficiency of our pretraining method by investigating how model performance is affected by subject diversity and sample volume (i.e., number of samples) in the training data used for fine-tuning.
To study the effect of subject diversity independent of sample volume, we separately varied the number of subjects and the number of samples in the training data.
The number of subjects was randomly sampled as $n_{subj} \in \{1, 2, 3, 4, 5\}$, and the number of samples was randomly sampled from those subjects as $n_{samples} \in \{50, 130, 340, 900, \text{all}\}$ (``all'' indicates that all available training samples were used).
Our sampling strategy selected sleep staging data without class stratification, and each sleep stage was represented at least once in the reduced datasets.
For each parameter combination, we trained the fully supervised and fine-tuned feature extractor configurations on all three datasets with three repetitions of 5-fold cross-validation.

\begin{figure}
    \centering
    \includegraphics[width=\textwidth]{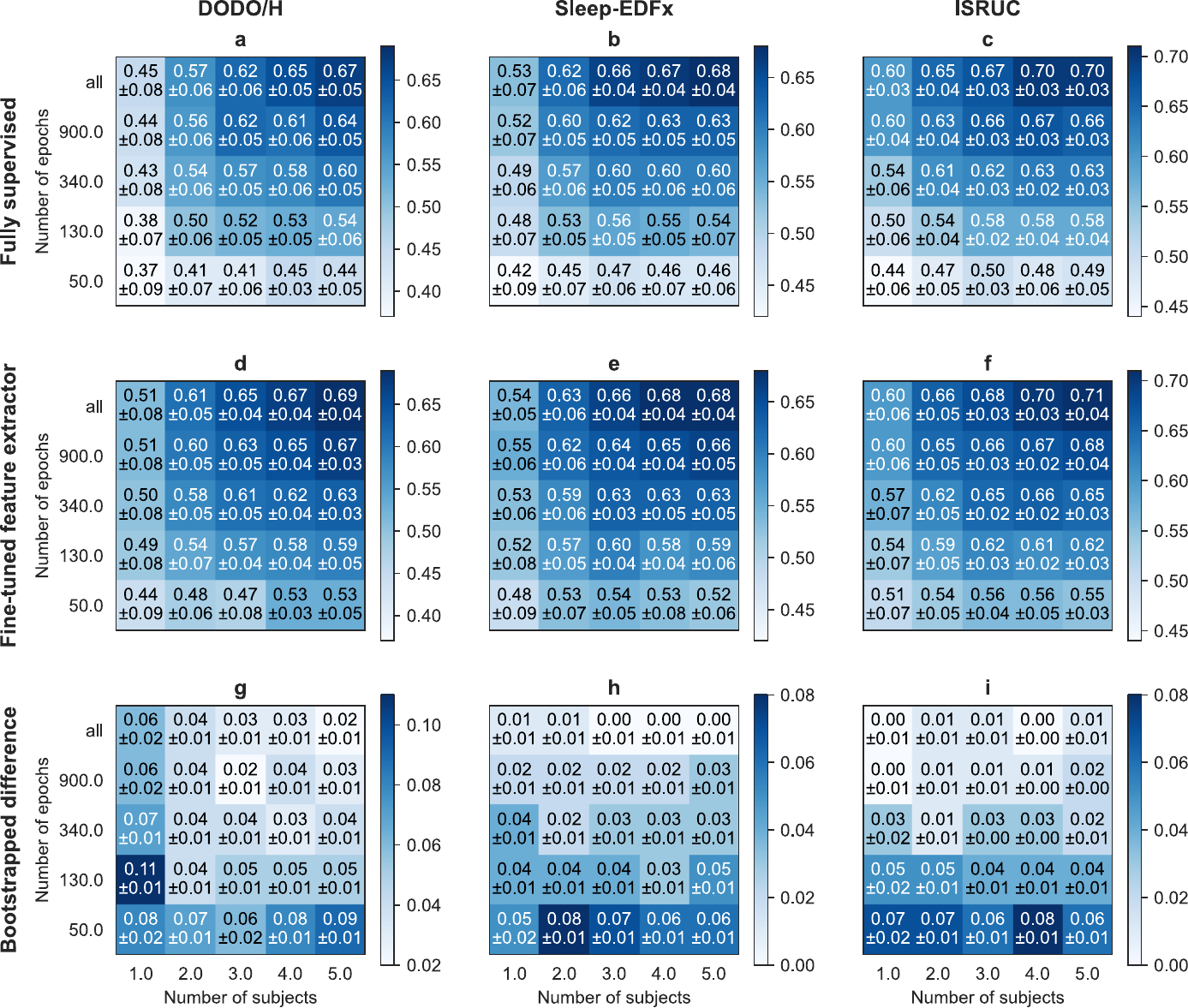}
    \caption{
        Detailed comparison of the fully supervised and fine-tuned feature extractor configurations in few-sample and few-subject regimes for each of the three datasets DODO/H (panels \textbf{a},\textbf{d},\textbf{g}), Sleep-EDFx (matrices \textbf{b},\textbf{e},\textbf{h}), and ISRUC (panels \textbf{c},\textbf{f},\textbf{i}).
        Panels (\textbf{a}--\textbf{c}) and Panels (\textbf{d}--\textbf{f}) show the macro F1 scores achieved by the fully supervised and the fine-tuned feature extractor configurations, respectively.
        The displayed scores are the mean and the standard deviation of the macro F1 scores measured on the test set over 15 trainings (3 repetitions of a 5-fold cross-validation).
      \linebreak Panels (\textbf{g--i}) show the average macro F1 scores achieved by the fine-tuned feature extractor configuration minus the average macro F1 scores achieved by the fully supervised configuration.
        These differences were calculated using a bootstrapping approach with 10,000 bootstrap samples.
        In this bootstrapping approach, we first paired the 15 macro F1 scores available for both configurations for each matrix entry based on the used seed and fold.
        We then calculated the difference between each pair of macro F1 scores.
        For each bootstrap sample, we sampled 15 of these differences with replacement and calculated their average.
        Finally, we display the mean and the standard deviation of these bootstrap samples.
    }
    \label{fig:subj-epoch-matrices}
\end{figure}

In our results, we observed that both the fully supervised and the fine-tuned feature extractor configurations benefited from increased subject diversity, even when the total number of training samples was held constant (see rows in Figure~\ref{fig:subj-epoch-matrices}a--f).
Similarly, both configurations benefited from an increased number of training samples when the number of subjects was held constant (see columns in Figure~\ref{fig:subj-epoch-matrices}a--f).
For the considered parameter ranges, the impact of reduced subject diversity and reduced sample volume on model performance appeared to be comparable.

Similar to the observations made in Section~\ref{sec:data-efficiency}, the fine-tuned feature extractor configuration achieved better macro F1 scores than the fully supervised configuration in the few-sample regime.
The performance gap between the two configurations was most evident when the number of samples was limited to 50, with bootstrap estimates of the mean differences in macro F1 scores ranging from 0.05 to 0.09 across datasets and subject numbers (see Figure~\ref{fig:subj-epoch-matrices}g--i).
These differences in macro F1 scores decreased as the number of samples increased.
When training on all available training samples, the fine-tuned feature extractor configuration achieving comparable or slightly better performance than the fully supervised configuration.
Interestingly, the performance gap between the two configurations with all available training samples was most pronounced for the DODO/H dataset.
The fine-tuned feature extractor configuration achieved average macro F1 scores that were 0.02--0.06 higher than those of the fully supervised configuration.
For the Sleep-EDFx and ISRUC datasets, the performance differences between the two configurations with all available training samples were minimal at 0.00--0.01.

Depending on the dataset, the fine-tuned feature extractor configuration showed varying degrees of improvement over the fully supervised configuration in the few-subject regime.
When training with data of a single subject, the fine-tuned feature extractor configuration achieved average macro F1 scores that were 0.06--0.11 higher across sample numbers than those of the fully supervised configuration for the DODO/H dataset (see rows in Figure~\ref{fig:subj-epoch-matrices}g).
This performance gap decreased as the number of subjects increased, with mean bootstrapped differences in macro F1 scores between 0.02 and 0.09 for five subjects.
In contrast, varying the number of subjects had less impact on the performance gap between the fine-tuned feature extractor and the fully supervised configurations for the Sleep-EDFx and ISRUC datasets (see rows in Figure~\ref{fig:subj-epoch-matrices}h--i).
The improvements achieved by the fine-tuned feature extractor configuration when trained with one subject (0.01--0.05 for the Sleep-EDFx dataset and 0.00--0.07 for the ISRUC dataset) were comparable to those achieved when trained with five subjects (0.00--0.06 for the Sleep-EDFx dataset and 0.01--0.06 for the ISRUC dataset).

\subsection{Priors Towards Frequency Information}
\label{sec:pretraining}

\begin{figure}[t]
    \centering
    \includegraphics[width=\textwidth]{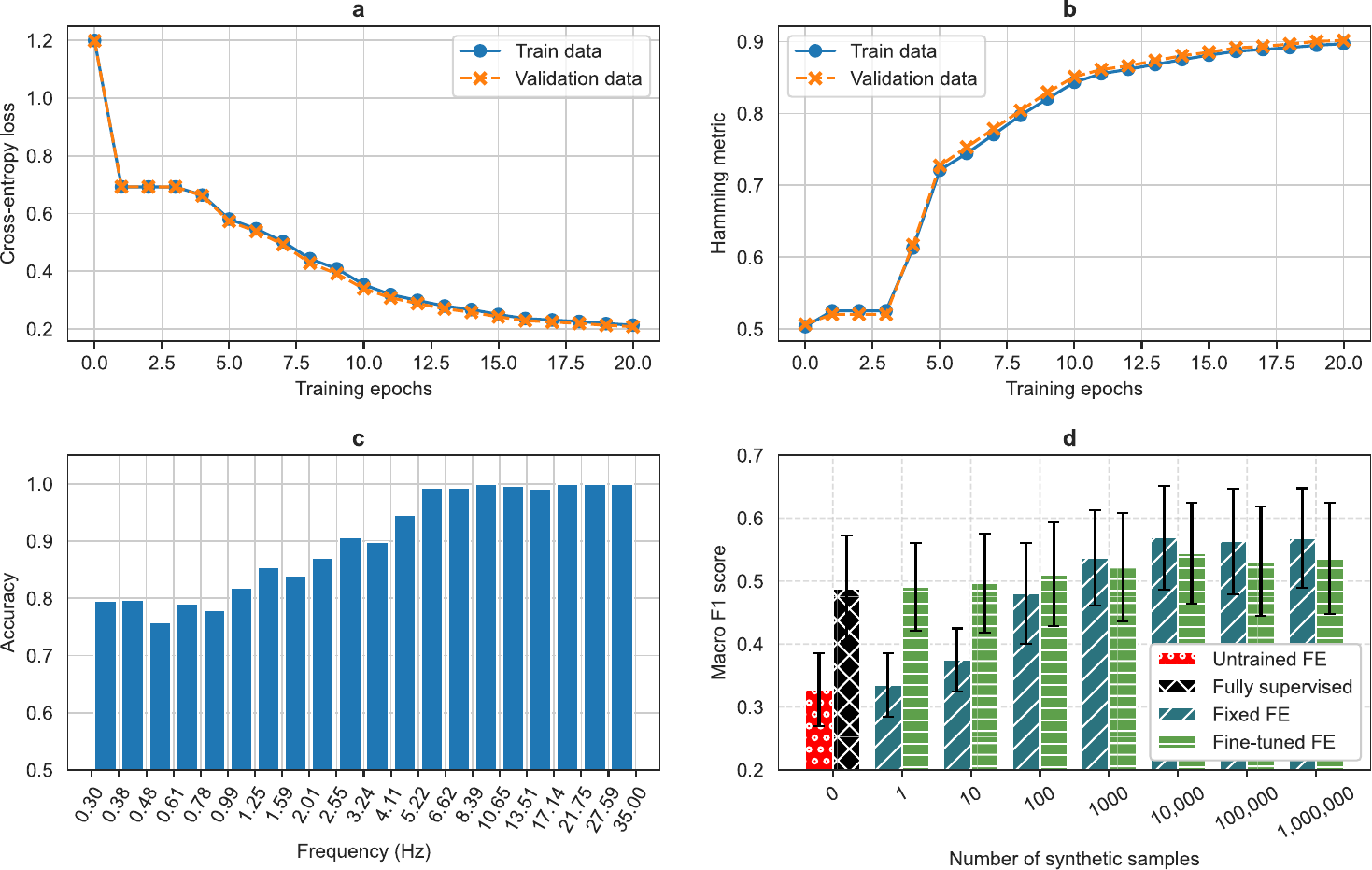}
    \caption{
            Analysis of the ``frequency pretraining'' task.
            Panel (\textbf{a}) shows the development of the loss function during a single pretraining run for both the training and validation data.
            The following two panels quantify the accuracy of a pretrained model to predict the frequencies of the synthetic signals.
            The hamming metric in panel (\textbf{b}) measures the overall accuracy summarized across all frequency bins, while panel (\textbf{c}) shows the accuracy for each frequency bin separately.
            Panel (\textbf{d}) illustrates how the diversity of the synthetic pretraining data affects the performance of the pretrained feature extractor (FE) after fine-tuning for sleep staging on a single subject from the DODO/H dataset.
            Each bar indicates the mean of the macro F1 scores on the validation data averaged over 3 repetitions of a 5-fold cross-validation, while the error bars show the standard deviation of the macro F1 scores.
    }
    \label{fig:pretraining-metrics}
\end{figure}

To get a better understanding of the pretraining process and the priors learned by the model, we recorded several metrics during pretraining.
The recorded loss function converged to a low value, indicating that the model has learned effectively (see Figure~\ref{fig:pretraining-metrics}a).
At the same time, the Hamming metric reached a high value of 0.9 on the validation data (see Figure~\ref{fig:pretraining-metrics}b).
This value can be interpreted as the model predicting the frequency bins that were used to create the synthetic signals with an accuracy of 90\%.
The model was particularly proficient in predicting higher frequencies starting from 2.5 Hz (\mbox{accuracies $>90\%$;} see Figure~\ref{fig:pretraining-metrics}c).
Lower frequencies, especially those below 1 Hz, were predicted with lower accuracy (75--85\%).

We hypothesize that the differences in prediction accuracy across frequency bins are not due to the pretrained model being unable to predict lower frequencies.
Instead, we believe this discrepancy arises from the varying width of the frequency bins (see x-axis of Figure~\ref{fig:pretraining-metrics}c) as the bins for lower frequencies were narrower than those for higher frequencies due to their logarithmic scaling.
While we applied a logarithmic binning scheme to increase our model's focus on the lower frequencies important for identifying slow wave sleep (N2 and N3 sleep), the narrow low frequency bands increase the difficulty of the pretraining task leading to decreased accuracies.
Preliminary exploration of this trade-off between model focus and task difficulty showed that the current binning scheme slightly outperforms a linear binning scheme in the fixed feature extractor setup.

Interestingly, the ability of the pretrained feature extractor to extract useful features for sleep staging was strongly influenced by the diversity of the synthetic pretraining data (see Figure~\ref{fig:pretraining-metrics}d).
We investigated this influence of sample diversity by pretraining models with varying amounts of different synthetic samples $n_{synthetic} \in \{1, 10, 100, 10^3, 10^4, 10^5, 10^6\}$.
To isolate the effect of sample diversity from the effect that the number of training steps has on model performance, we kept the number of gradient updates per training epoch constant by under- or oversampling the synthetic data to 100,000 samples.
The pretrained models were then trained for sleep staging using data from one random subject from the DODO/H dataset and evaluated on the validation data of each cross-validation fold.
As in the previous experiments, we performed three repetitions of a 5-fold cross-validation scheme for each number of synthetic samples.

Both the fixed feature extractor and the fine-tuned feature extractor configurations required a high level of diversity in the synthetic pretraining data to achieve good sleep staging performance (see Figure~\ref{fig:pretraining-metrics}d).
When pretrained with only one synthetic sample, the performance of the two training configurations with pretraining differed only slightly from the performance of the configurations without pretraining (untrained feature extractor and fully supervised configurations).
As the number of synthetic samples was increased to more than 100, the performance of the pretrained models improved substantially until reaching a plateau at around 10,000 samples.
We hypothesize that this plateau was reached because the model had learned all the relevant features from the pretraining task, and additional samples did not provide any further benefit.
The simplicity of the pretraining task could also explain the negligible performance differences between the pretrained configurations with less than 100 synthetic samples and the training configurations without pretraining.
For such few synthetic samples, the model may have memorized the synthetic data rather than learn general features useful for sleep staging.

\subsection{Comparison to Self-Supervised Methods}

We compared our frequency pretraining approach (FPT) with two popular self-supervised learning (SSL) methods, SimCLR~\cite{Chen2020a} and VICReg~\cite{Bardes2022}, using the same model architecture as in FPT.
The projection head (SimCLR) and expander (VICReg) followed their original designs~\cite{Chen2020a,Bardes2022}.
Different from the conceptually simple multi-label classification task in FPT, both SSL methods follow the objective of contracting representations of data-augmented ``positive'' samples while keeping representations of dissimilar ``negative'' samples far apart (see original works for details~\cite{Chen2020a,Bardes2022}).
Positive samples were generated using standard data augmentations: amplitude scaling (random factor 0.5--2), Gaussian noise injection ($\sigma$ = 0.05), random temporal masking (10 segments of 1.5--3 s), time shifting (up to $\pm$1.5 s), and time warping (random factor 0.67--1.5).
We evaluated two pretraining configurations: pretraining on (i) synthetic data (as in FPT) and (ii) EEG recordings from the ISRUC and Sleep-EDFx datasets.
Fine-tuning was performed on DODO/H datasets under low-data (50 random epochs from one random subject) and high-data (all available training data) regimes, respectively.

When pretrained on synthetic data, both SimCLR and VICReg achieved performance comparable to the FPT method, with FPT slightly outperforming the SSL methods in the high-data regime when the feature extractor was fine-tuned (average MF1 scores of 0.80 versus 0.81, see Table~\ref{tab:comparison_ssl}).
Pretraining on EEG data yielded only modest improvements, with both SSL methods showing slightly better performance in the low-data regime when the feature extractor was fixed (difference in average MF1 scores of 0.02) and in the high-data regime when fine-tuning the feature extractor (difference in average MF1 scores of 0.01).
Compared to the EEG-pretrained SimCLR and VICReg models in the high-data regime, FPT showed similar performance (average MF1 scores of 0.76 and 0.81) but, unlike the two SSL methods, did not require EEG data for pretraining.
In the low-data regime, the SSL methods performed comparable to the FPT method when fine-tuning the feature extractor and achieved slightly higher scores than FPT when keeping the feature extractor fixed (average MF1 scores of 0.44 and 0.45 versus 0.42).
None of the differences in MF1 scores between the SSL methods and FPT were significant according to a two-sided Wilcoxon signed-rank test (\emph{p}-values $>$ 0.05).

\setlength{\tabcolsep}{10pt} % Adjust column separation

\begin{table}[h]
    \centering
	\caption{
		Comparison of different pretraining methods when fine-tuned with 50 random sleep epochs of one subject (low-data regime) or the full training set (high-data regime) of the DODO/H dataset.
        During fine-tuning, the feature extractor was either kept fixed (FE fixed) or updated together with the classifier (FE fine-tuned).
        We pretrained SimCLR and VICReg on the same synthetic data used for the frequency pretraining task, in addition to the traditional approach of pretraining on EEG data (i.e., ISRUC and Sleep-EDFx datasets).
        MF1 scores were calculated as the average of three repetitions of a five-fold cross validation (standard deviation in parentheses).
		\label{tab:comparison_ssl}
	}
		\begin{tabular}{ccccc}
		\toprule
		\textbf{Pretraining Method}	& \multicolumn{2}{c}{\textbf{Avg. MF1 Low-Data Regime}}	& \multicolumn{02}{c}{\textbf{Avg. MF1 High-Data Regime}}\\
        \cmidrule{2-5}
		& \textbf{FE Fixed}                                       & \textbf{FE Fine-Tuned}                                    & \textbf{FE Fixed} & \textbf{FE Fine-Tuned} \\
		\midrule
		SimCLR, pretrained with synth. data & 0.42 (0.08) & 0.44 (0.09) & 0.76 (0.02) & 0.80 (0.02) \\
 SimCLR, pretrained with EEG data    & 0.44 (0.08) & 0.43 (0.09) & 0.76 (0.02) & 0.81 (0.02) \\
 VICReg, pretrained with synth. data & 0.43 (0.07) & 0.44 (0.08) & 0.76 (0.02) & 0.80 (0.02) \\
 VICReg, pretrained with EEG data    & 0.45 (0.08) & 0.45 (0.09) & 0.76 (0.02) & 0.81 (0.01) \\
 Frequency pretraining               & 0.42 (0.06)                          & 0.44 (0.09)                          & 0.76 (0.01)                          & 0.81 (0.02)                          \\
		\bottomrule
		\end{tabular}
\end{table}

\section{Discussion}

In this work, we propose a novel pretraining scheme for EEG time series data that leverages synthetic data generated by a simple random process.
We specialized the hyperparameters of the pretraining task to the typical frequency range and distribution of sleep EEG signals and demonstrated the effectiveness of this task for sleep stage classification.
Due to the availability of several open sleep staging datasets~\cite{Obeid2016,Zhang2018}, we were able to fully control the amount and diversity of the training data, which allowed us to study the impact of our method in different data regimes.
We hypothesize that our pretraining scheme could be particularly beneficial in few-sample and few-subject regimes, which we argue could benefit greatly from the priors towards frequency information that a model learns during pretraining.

Our results confirm the effectiveness of our pretraining scheme, particularly in few-sample and few-subject regimes.
Pretrained models outperformed non-pretrained models when fine-tuned with a reduced number of subjects or training samples (see rows and columns in Figure~\ref{fig:subj-epoch-matrices}g--i, respectively).
The performance gap between pretrained and non-pretrained models was most pronounced in the few-sample regime, where pretrained models consistently achieved improvements over non-pretrained models across multiple datasets.
In the few-subject regime, this performance gap was not as consistent across datasets, with our pretraining method showing the most substantial improvements in the DODO/H dataset.
Our findings support observations made in the field of Self-Supervised Learning (SSL) that pretrained models generally have better data efficiency than fully supervised ones~\cite{Banville2021,Eldele2023}.
In contrast to SSL methods, however, our pretraining scheme improves data efficiency without requiring empirical data, while achieving comparable or only slightly reduced performance (see Table~\ref{tab:comparison_ssl}).
Interestingly, we observed that two data-augmentation-based SSL methods performed well even when pretrained with synthetic data instead of EEG data (see Table~\ref{tab:comparison_ssl}), suggesting that SSL approaches are promising, yet computationally intensive, alternatives to the FPT method.
Further exploration of the use of SSL methods for pretraining on synthetic data is warranted and seems promising.
Generating synthetic data can be much cheaper and more cost-effective than collecting EEG data, as generating 10,000 samples of the synthetic data used in this study required only $8.17 \pm 0.37$ s of a single consumer-grade CPU core (Lenovo Legion S7 16IAH7 Laptop with an Intel\textsuperscript{\circledR} Core\textsuperscript{\texttrademark} i7-12700H CPU).

We hypothesize that the potential of synthetic data stems from the priors that the model learns during pretraining.
These priors could prevent overfitting to a small number of training samples, particularly those from minority classes (e.g., N1 sleep), or subject-specific features, which is especially problematic in situations with very little training data.
As expected, we observed that all of our training configurations improved with a larger training dataset (see Figure~\ref{fig:subj-vs-mf1}).
This aligns with the prevalent view in the literature that deep learning models for sleep staging need substantial amounts of diverse data to perform well~\cite{Phan2022, AlvarezEstevez2023, Fiorillo2019, Fiorillo2023}.
When trained with the full training data, pretrained models performed comparably to fully supervised models (see Figure~\ref{fig:subj-vs-mf1}), achieving macro F1 scores similar to those of other deep learning approaches for sleep staging~\cite{Phan2022, Gaiduk2023}.
In conclusion, our pretraining method was most beneficial in situations with limited training data, where it outperformed models trained from scratch, but had less impact in situations with large amounts of training data.

We further observed that, while the frequency content of a signal is crucial for sleep staging, deep neural networks extract additional information from the data that exceeds the frequency domain.
The importance of the frequency content of a signal for sleep staging is demonstrated by the high macro F1 scores achieved by the fixed feature extractor configuration and the substantial performance improvements it achieved over untrained feature extractors (see Figure~\ref{fig:subj-vs-mf1}).
We attribute the performance gap between the two training configurations to the priors learned during pretraining the feature extractor.
These priors biased the model to extract frequency information from the data, which it achieved with high accuracy after pretraining (see Figure~\ref{fig:pretraining-metrics}b,c).
Our finding is consistent with previous studies that reported frequency-based features to be important for sleep staging~\cite{Motamedi2014}.
When the feature extractor of our model was allowed to be fine-tuned after pretraining, model performance increased (see Figure~\ref{fig:subj-vs-mf1}).
We hypothesize that this increase in the macro F1 score is due to the feature extractor learning to extract information beyond the frequency content of the signal during fine-tuning.
This hypothesis is in line with the AASM annotation guidelines~\cite{Berry2020}, which consider several frequency-unrelated features essential for sleep staging.
These features include time-domain information, which is important for spindles as well as amplitudes and specific patterns, such as k-Complexes~\cite{Fiorillo2019}.
Recent studies that applied feature engineering approaches to sleep staging further support our hypothesis by including additional features from the time-domain in the models used~\cite{Vallat2021,VanDerDonckt2023}.
In their work, Vallat and Walker analyzed the most important features for their model and found that time-related features, such as the time elapsed from the beginning of a recording, were among the top 20 most important features~\cite{Vallat2021}.
Although it remains unclear what additional information our pretrained models learn during fine-tuning, our method offers a promising avenue for future research into the interpretability of deep neural networks for sleep staging.

There are several opportunities for future work that could build upon our findings.
One promising direction is to explore the pretraining task in more detail, for example, by investigating the synthetic data generation process and the impact of changing the used frequency range.
Similar to previous work in the vision domain~\cite{Baradad2021}, it could also be promising to investigate which structural properties of synthetic time series are important for sleep staging.
This could be achieved by defining new pretraining tasks that are based on different data generation processes that incorporate more complex structures like desynchronized phases across channels, noise, or polymorphic amplitude variations.
Exploring such data generation processes may lead to a better understanding of what constitutes ``natural'' EEG time series and what information, besides the frequency content, is essential for sleep staging.
In addition, we suggest exploring models with greater capacity and less inductive bias than the CNN-based architecture used in this work, such as transformer models~\cite{Vaswani2017,Brandmayr2022}, which we expect to benefit even more from our pretraining method.
Pretraining such models with synthetic data may alleviate their need for large amounts of training data~\cite{Dosovitskiy2021}.
Another avenue for future research is to investigate whether our pretraining method is beneficial for specific cohorts of subjects, such as patients with a specific disorder or specific age groups.
Although we did investigate datasets with different demographics in this work, we did not perform detailed analyses of the impact of these demographics on model performance.
Finally, it could be insightful to compare our approach with a broader range of SSL methods~\cite{Liu2023} and data augmentation strategies that employ synthetic EEG generators~\cite{Lashgari2020,Habashi2023}.
To enable such comparisons and to facilitate future research in this direction, we make our code available online~\cite{Grieger2024}.

Our method presents a novel solution to address important issues that affect current deep learning models in the EEG time series domain, without requiring large amounts of patient data.
We expect our approach to be advantageous in various applications where EEG data is scarce or derived from a limited number of subjects, such as brain--computer interfaces~\cite{Ko2021} or neurological disorder detection~\cite{vanDijk2022}.

\section{Materials and Methods}

\label{sec:methods}

We trained deep neural networks in two phases: a pretraining phase based on synthetic data and a fine-tuning phase based on clinical sleep staging data.
The training process is illustrated in Figure~\ref{fig:pretraining-scheme}.

\phantomsection
\subsection{Phase 1: Pretraining with Synthetic Data}

In the pretraining phase, we trained a deep neural network to predict the frequencies of randomly-generated synthetic time series (``frequency pretraining'')~\cite{Grieger2024a}.
The deep neural network consisted of a feature extractor to produce features and a classifier to predict which frequencies are present in the synthetic time series.
Instead of training the model to predict the exact frequencies, we defined frequency ranges (frequency bins) and let the model identify the bins containing frequencies present in the signal (therefore turning the problem into a multi-label classification task).

\phantomsection
\subsubsection{Synthetic Data Generation}
The synthetic samples used for pretraining consisted of a time series signal \linebreak $s(t) \in \mathbb{R}^{3 \times 3000}$, featuring 3 channels of 30 s sampled at 100 Hz, and an associated label vector $l$.
The synthetic time series were created by summing sine waves with randomly sampled frequencies, while the label vector encoded from which frequency ranges (frequency bins) the frequencies were sampled.
Since we wanted to test our method in the context of sleep staging, we followed the American Academy of Sleep Medicine (AASM) guidelines and focused on the spectral band between 0.3--35 Hz.
This range was divided into 20 frequency bins on a base-2 logarithmic scale, resulting in a label vector $l \in \{0,1\}^{20}$ with binary entries for each bin that indicated whether frequencies in a given bin were present (1) or absent (0) in the time series.
To generate a synthetic time series, we first randomly selected a subset of the 20 frequency bins, setting the corresponding entries in the label vector $l$ to 1 (each bin was selected with a probability of 50\%).
Next, we independently sampled frequencies from the selected bins for each channel and generated sine waves with these frequencies.
The sine waves were then shifted by a random phase sampled from a uniform distribution between 0 and $2\pi$, summed, and normalized to create the synthetic signal. Therefore, one channel $s_c(t)$ of a synthetic signal $s(t) = (s_1(t), s_2(t), s_3(t))$ is defined as
\begin{equation}
    s_c(t) = \frac{\tilde{s}_c(t)-\mu_c}{\sigma_c}, \quad  \tilde{s}_c(t) = \sum_i^{n_f} I_i\, \sin(2 \pi \,t f_{c,i} + \phi_i), \quad I_i \in \mathcal{U}\{0,1\},
\end{equation}
where $t$ denotes time (in seconds), $n_f$ denotes the number of frequency bins, $I_i$ denotes the indicator function encoding whether a frequency within a bin $i$ is present (1) or not (0), $f_{c,i}$ denotes the frequencies of the sine waves (in Hertz), $\phi_i$ denotes their phase (in radians), and $\mu_c$ and $\sigma_c$ denote mean and standard deviation of $\tilde{s}_c$.
Note that the phases $\phi_i$ were sampled for each frequency bin independently but remained the same across channels to keep the data generation process simple, while the frequencies $f_{c,i}$ were sampled independently for each bin and channel.

For each training run, we generated 101,000 synthetic samples, using 100,000 samples for training and the remaining 1000 samples as a validation set to track various metrics.
The number of synthetic samples, frequency bins, and the logarithmic scale that defines the boundaries between the bins were determined through preliminary experiments where we explored a wide range of hyperparameters.

\subsubsection{Model}

We based our model on the TinySleepNet architecture, a conceptually simple deep neural network for sleep staging that has previously demonstrated competitive results~\cite{Supratak2020}.
This architecture consists of a convolutional feature extractor that extracts features from individual sleep epochs and a classifier that aggregates these feature across multiple epochs to perform sleep staging.
The feature extractor consisted of four convolutional layers with 128 filters each, a kernel size of 50, 8, 8, and 8, respectively, and a stride of 25, 1, 1, and 1, respectively.
Following each convolutional layer was a batch normalization layer and a ReLU activation function.
The outputs of the first and last convolutional layer were passed through max pooling layers that reduced the temporal dimension of the feature maps by a factor of 8 and 4, respectively.
After both max pooling layers, we applied dropout with a dropout rate of 0.5.

The feature extractor was followed by a classifier, which consisted of a dense layer with 80 neurons and ReLU activation function as well as a dense layer with 20 neurons and sigmoid activation function (i.e., one output for each frequency bin).
We defined a threshold of 0.5 to determine whether a frequency from a frequency bin was present in the signal (output greater than 0.5) or not (output less than or equal to 0.5).

\subsubsection{Pretraining for Frequency Prediction}

We pretrained models for 20 training epochs using the Adam optimizer~\cite{Kingma2015}, a fixed learning rate of $10^{-4}$, a batch size of 64, and a binary cross-entropy loss.
This loss function is commonly employed for multi-label classification problems with one-hot encoded labels.
It is defined as
\begin{equation}
    L_{bce} = - \frac{1}{N} \frac{1}{n_f} \sum_{i=1}^{N} \sum_{j=1}^{n_f} y_{i,j} \cdot \log(\hat{y}_{i,j}) + (1 - y_{i,j}) \cdot \log(1 - \hat{y}_{i,j}),
\end{equation}
where $N$ is the number of samples, $n_f$ is the number of frequency bins, $y_{i,j}$ is the true one-hot encoded label of sample $i$ and frequency bin $j$, and $\hat{y}_{i,j}$ is the predicted one-hot encoded label of sample $i$ and frequency bin $j$.

In addition to the loss function, we recorded the hamming metric on the training and validation data after each training epoch.
This metric is derived from the hamming loss, a common metric for multi-label classification problems~\cite{Sokolova2009}.
Specifically, the hamming metric tracks the fraction of correctly predicted frequency bins and is defined as
\begin{equation}
    H = \frac{1}{N} \frac{1}{n_f} \sum_{i=1}^{N} \sum_{j=1}^{n_f} I_{\{y_{i,j} = \hat{y}_{i,j}\}},
\end{equation}
where $I$ is the indicator function, which assumes the value 1 if the true one-hot encoded label $y_{i,j}$ of sample $i$ and frequency bin $j$ is equal to the predicted one-hot encoded label $\hat{y}_{i,j}$ of sample $i$ and frequency bin $j$.

\subsection{Phase 2: Fine-Tuning with Sleep Staging Data}

In the fine-tuning phase, we used the pretrained feature extractor of the pretraining phase for sleep stage classification on EEG and EOG data.
Sleep stage classification is a multi-class classification problem where a short segment of EEG and EOG data (sleep epoch) is assigned to one of five sleep stages (Wake, N1, N2, N3, REM) based on the underlying brain activity.
In our approach, the feature extractor of our model produced features from individual sleep epochs of EEG and EOG data.
The extracted features were then concatenated over a sequence of multiple epochs and aggregated by a classifier to predict the sleep stage of the middle epoch in the input sequence (see Figure~\ref{fig:pretraining-scheme}).
By incorporating the surrounding sleep epochs into the classifier's input, we provided additional temporal context, potentially enhancing the accuracy of sleep stage prediction.

\phantomsection
\subsubsection{Sleep Staging Data}
\label{sec:sleep-staging-data}
To test our approach, we used four publicly available datasets encompassing a diverse range of subjects, including both healthy individuals and those with various medical conditions such as sleep apnea, REM sleep behavior disorder, and affective disorders: DODO~\cite{Guillot2020}, DODH~\cite{Guillot2020}, Sleep-EDFx~\cite{Kemp2000,Mourtazaev1995}, and ISRUC~\cite{Khalighi2016}.
The DODO and DODH datasets contain 55 recordings from subjects diagnosed with obstructive sleep apnea (OSA) and 25 recordings from healthy subjects, respectively.
We combined the two datasets into a single dataset with 80 recordings, which we refer to as DODO/H.
Each recording stems from a different subject and was split into 30-s non-overlapping windows (sleep epochs), which were annotated with sleep stages (Wake, N1, N2, N3, REM) by five sleep experts following the AASM guidelines~\cite{Berry2020}.
We aggregated these five expert annotations into a consensus label for each sleep epoch using majority voting, with ties resolved by selecting the sleep stage determined by the most reliable scorer (defined as the scorer with the highest average agreement with all other scorers)~\cite{Guillot2020}.
Epochs that the consensus of the sleep experts annotated as artifacts were removed from the datasets.
In our experiments, we focused on three EEG and EOG derivations available in both datasets and recommended by the AASM guidelines~\cite{Berry2020}: C3-M2, F3-M2, EOG1.
All channels were sampled at 250~Hz.

The Sleep-EDFx dataset is provided on the PhysioNet platform~\cite{Goldberger2000} and contains \linebreak 197 recordings from two studies~\cite{Kemp2000,Mourtazaev1995}, of which we used the 153 recordings of the ``Sleep Cassette'' study~\cite{Mourtazaev1995}.
The recordings stem from 78 healthy subjects with two recordings per subject, except for subjects 13, 36, and 52, who have only one recording each due to technical issues.
The recordings were split into 30-s sleep epochs and annotated with sleep stages (Wake, N1, N2, N3, N4, REM) according to the 1968 Rechtschaffen and Kales manual~\cite{Rechtschaffen1968} by a single expert per recording.
To be consistent with the annotations of the other datasets, we combined the N3 and N4 stages into a single N3 stage and removed data annotated as artifacts or movement.
Furthermore, we removed excess daytime data from the recordings by cropping them to begin 30 min before the first annotated non-Wake stage and end 30 min after the last annotated non-Wake stage.
All available EEG and EOG channels (Fpz-Cz, Pz-Oz, EOG horizontal) were sampled at 100~Hz.

The ISRUC dataset~\cite{Khalighi2016} contains 126 recordings and is divided into three subgroups.
Subgroup one contains 100 recordings from 100 subjects suffering from various disorders affecting sleep (including OSA, REM disorder, affective disorder, snoring), subgroup two contains 16 recordings from 8 subjects diagnosed with OSA and snoring (2 recordings per subject), and subgroup three contains 10 recordings from 10 healthy subjects.
Due to technical issues, we excluded two recordings (recordings 8 and 100) from the first subgroup.
All recordings were split into 30-s sleep epochs and annotated with sleep stages by two experts following the AASM guidelines~\cite{Berry2020}.
We only used the annotations of the first expert and focused on the EEG and EOG channels C3-A2, F3-A2, and LOC-A2 sampled at 200~Hz.

To prepare the data for training, we filtered the signals between 0.3 and 35 Hz with an 8th-order zero-phase Butterworth filter and downsampled the filtered signals to 100 Hz using polyphase filtering if necessary.
We then normalized the amplitudes of each individual sleep epoch by subtracting the median and dividing the result by the interquartile range of its amplitude distribution.
Finally, we clipped all signal amplitudes above 20 or below $-$20 to minimize outliers.

After preprocessing, the three datasets (DODO/H, Sleep-EDFx, ISRUC) were split subject-wise into five folds for cross-validation.
Note that we created a separate cross-validation split for each dataset, resulting in three different 5-fold cross-validation configurations.
When splitting the DODO/H and the ISRUC datasets, we ensured that the recordings within each split were balanced in regard to the sub-datasets or subgroups.
In addition to splitting the data into training and test folds, we reserved approximately 10\% of the data from each fold as validation data for hyperparameter tuning and early stopping (i.e., one recording of DODO and DODH each, two subjects of Sleep-EDFx, and two recordings from the first subgroup of ISRUC for each fold).
We validated our models on the validation data of the four training folds.
The validation data of the test fold was kept separate and not involved in validation or testing to prevent data leakage.

\subsubsection{Model}

Since the sleep epochs used for fine-tuning had the same format as the synthetic data generated for pretraining (i.e., 3 channels with 30 s of data sampled at 100 Hz), we kept the architecture of the feature extractor unchanged.
However, we replaced the classifier with a different architecture to predict sleep stages instead of frequency bins.
The classifier was inspired by the TinySleepNet architecture~\cite{Supratak2020} and consisted of a bidirectional Long-Short Term Memory (LSTM)~\cite{Hochreiter1997} layer with a hidden size of 128, a dropout layer with a dropout rate of 0.5, and a dense layer with 5 neurons and softmax activation function.

\subsubsection{Fine-Tuning for Sleep Stage Classification}

We fine-tuned the models with training samples consisting of 11 sleep epochs that provided the classifier with context information from 5 epochs before and after the epoch to be classified.
To ensure that the model could use the same aggregation process for all sleep epochs of a recording, we padded the first and last five sequences of each recording with zeros to the full sequence length.
Fine-tuning was performed using the Adam optimizer~\cite{Kingma2015}, a fixed learning rate of $10^{-4}$, weight decay of $10^{-3}$, a batch size of 32, and the categorical cross-entropy loss.
To prevent overfitting, we limited each training run to a maximum of 50 training epochs and stopped training early if the macro F1 score on the validation data did not improve for 10 training epochs (early stopping).
Additionally, we clipped all gradients with a maximum norm greater than 5.0 to prevent exploding gradients.

We tracked model performance by recording the macro F1 score on the training and validation data after each training epoch.
The macro F1 score is calculated as the average F1 score~\cite{Tharwat2021} across all sleep stages and is defined as
\begin{equation}
    \overline{F_1} = \frac{1}{k} \sum_{i=1}^{k} 2 \cdot \frac{p_i \cdot r_i}{p_i + r_i},
\end{equation}
where $k$ is the number of sleep stages, $p_i$ is the precision of sleep stage $i$, and $r_i$ is the recall of sleep stage $i$.

In all four training configurations (Fully Supervised, Fixed Feature Extractor, Fine-Tuned Feature Extractor, and Untrained Feature Extractor), we performed training using the cross-validation scheme described in Section~\ref{sec:sleep-staging-data}.
A single training run in this cross-validation scheme included both pretraining and fine-tuning where applicable.
If not specified otherwise, all reported results were obtained on the test folds of the cross-validation splits, while the small part of the training data reserved for validation was used as early stopping set during fine-tuning.

To ensure comparability between different training and data configurations, we seeded the model initialization, the synthetic data generation during pretraining, and the data subsampling when fine-tuning with a limited amount of training data.
This strategy reduced the influence of random initialization and data sampling on comparisons between training configurations.
In particular, seeding ensured that models were fine-tuned with the same data across all training configurations in the few-sample and few-subject regimes.
Furthermore, when training with a reduced amount of data, we maintained a constant number of gradient updates per training epoch by duplicating each training sample
$\lfloor \frac{N}{N_{red}} \rfloor$ times, where $N$ is the number of samples in the full training data and $N_{red}$ is the number of samples after reduction.

Experiments were performed on an NVIDIA DGX A100 workstation with eight NVIDIA A100 GPUs and a Dell workstation with a single NVIDIA RTX A6000 GPU.

\vspace{6pt}

\section*{Author Contributions}
Conceptualization, N.G. and S.B.; methodology, N.G.; software, N.G.; validation, N.G.; formal analysis, N.G.; investigation, N.G., S.M. and S.B.; resources, S.B.; data curation, N.G.; writing---original draft preparation, N.G. and S.B.; writing---review and editing, N.G., S.M. and S.B.; visualization, N.G. and S.B.; supervision, S.M. and S.B.; project administration, N.G. and S.B. All authors have read and agreed to the published version of the manuscript.

\section*{Funding}
This research received no external funding.

\section*{Data Availability}
The DODO and DODH datasets analyzed in the current study are described in Guillot~et~al.~\cite{Guillot2020} and are publicly available (see \url{https://github.com/Dreem-Organization/dreem-learning-open} (accessed on 30 March 2023) for how to access the datasets).
The Sleep-EDFx dataset is described in Kemp~et~al.~\cite{Kemp2000} and Mourtazaev~et~al.~\cite{Mourtazaev1995} and is publicly available at \url{https://physionet.org/content/sleep-edfx/1.0.0/} (accessed on 30 March 2023).
The ISRUC dataset is described in Khalighi~et~al.~\cite{Khalighi2016} and is publicly available at \url{https://sleeptight.isr.uc.pt/}(accessed on 13 April 2023).
The source code necessary to reproduce our results is available on GitHub at \url{https://github.com/dslaborg/frequency-pretraining/tree/full-paper-version} (accessed on 29 August 2025).

\section*{Acknowledgments}
We are grateful to M. Reißel and V. Sander for providing us with computing resources.

\section*{Conflicts of Interest}
The authors declare no conflicts of interest.

\bibliographystyle{abbrv}

\end{document}